\documentclass[journal]{IEEEtran}
\usepackage{graphicx}
\usepackage{comment}
\usepackage{amsmath,amssymb} 
\usepackage{color}
\usepackage{xcolor}
\usepackage{multirow}
\usepackage{ulem}
\usepackage{bm}
\usepackage{cite}

\ifCLASSINFOpdf
\else
\fi

\begin{document}
\title{Joint Hand-object 3D Reconstruction from a Single Image with Cross-branch Feature Fusion}


\author{Yujin Chen$^*$,
        Zhigang Tu$^*$,~\IEEEmembership{Member,~IEEE,}
        Di Kang,
        Ruizhi~Chen,~\IEEEmembership{Member,~IEEE,}
        Linchao~Bao$^\dag$,
        Zhengyou~Zhang,~\IEEEmembership{Fellow,~IEEE}
        and~Junsong~Yuan,~\IEEEmembership{Fellow,~IEEE,}
\thanks{Y. Chen, Z. Tu and R. Chen are with the State Key Laboratory of Information Engineering in Surveying, Mapping and Remote Sensing, Wuhan University, Wuhan 430079, China. (e-mail:\{yujin.chen, tuzhigang, ruizhi.chen\}@whu.edu.cn).}
\thanks{D. Kang, L. Bao and Z. Zhang are with Tencent AI Lab, Shenzhen
518057, China (e-mail:\{dkang, linchaobao, zhengyou\}@tencent.com).}
\thanks{J. Yuan is with the Computer Science and Engineering Department,
University at Buffalo, Buffalo, NY 14228, USA. (email: jsyuan@buffalo.edu).}
\thanks{$^*$Equal contributions.}
\thanks{$^\dag$L. Bao is the corresponding author.}
}

%
%

\markboth{
}%
{Shell \MakeLowercase{\textit{et al.}}: Bare Demo of IEEEtran.cls for IEEE Journals}
%



\maketitle

\begin{abstract}
Accurate 3D reconstruction of the hand and object shape from a hand-object image is important for understanding human-object interaction as well as human daily activities.
Different from bare hand pose estimation, hand-object interaction poses a strong constraint on both the hand and its manipulated object, which suggests that hand configuration may be crucial contextual information for the object, and vice versa.
However, current approaches address this task by training a two-branch network to reconstruct the hand and object separately with little communication between the two branches.
In this work, we propose to consider hand and object jointly in feature space and explore the reciprocity of the two branches.
We extensively investigate cross-branch feature fusion architectures with MLP or LSTM units.
Among the investigated architectures, a variant with LSTM units that enhances object feature with hand feature shows the best performance gain.
Moreover, we employ an auxiliary depth estimation module to augment the input RGB image with the estimated depth map, which further improves the reconstruction accuracy.
Experiments conducted on public datasets demonstrate that our approach significantly outperforms existing approaches in terms of the reconstruction accuracy of objects.
\end{abstract}


%
\IEEEpeerreviewmaketitle

\section{Introduction}
There has been a growing interest in analyzing human hands from images, due to a wide range of immersive applications, e.g., virtual reality (VR), augmented reality (AR) and human-computer interaction \cite{holl2018efficient,tu2018semantic,tu2019action}.
Significant progresses on hand detection~\cite{bambach2015lending,deng2017joint,kolsch2004robust,morerio2013hand,Narasimhaswamy_2019_ICCV} and hand pose estimation \cite{lugaresi2019mediapipe,mueller2018ganerated,spurr2018cross,yuan2018depth,zimmermann2017learning} have been witnessed.
However, most of the existing studies assume bare hands, which is often not the case when hands are interacting with some objects.
Recent research began to address the problem by capturing 3D relationships between hands and objects, which can facilitate better  understanding of hand-object interactions in hand-related motion recognition, human action interpretation, robotic behavior imitation, etc.
\begin{figure}[t]
	\begin{center}
		\includegraphics[width=1\linewidth]{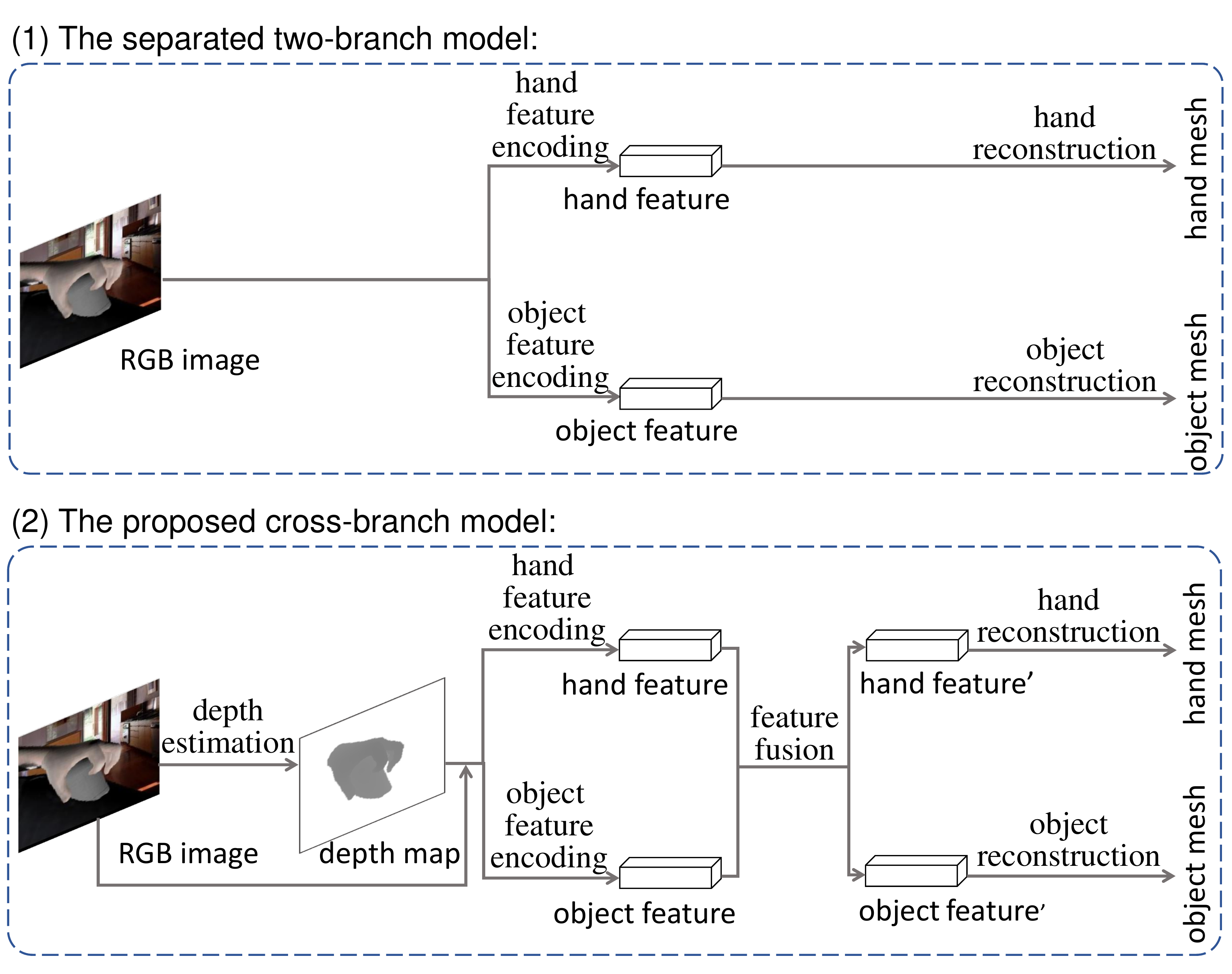}
	\end{center}
	\vspace{-3mm}
	\caption{Overview of the separated two-branch model \cite{hasson2019learning} and the proposed cross-branch model for joint hand-object reconstruction.}
	\label{fig:1}
	\vspace{-2mm}
\end{figure}

\begin{figure*}[t]
\centering
\makebox[0pt][c]{\parbox{1\textwidth}{%
		\includegraphics[width=1\linewidth]{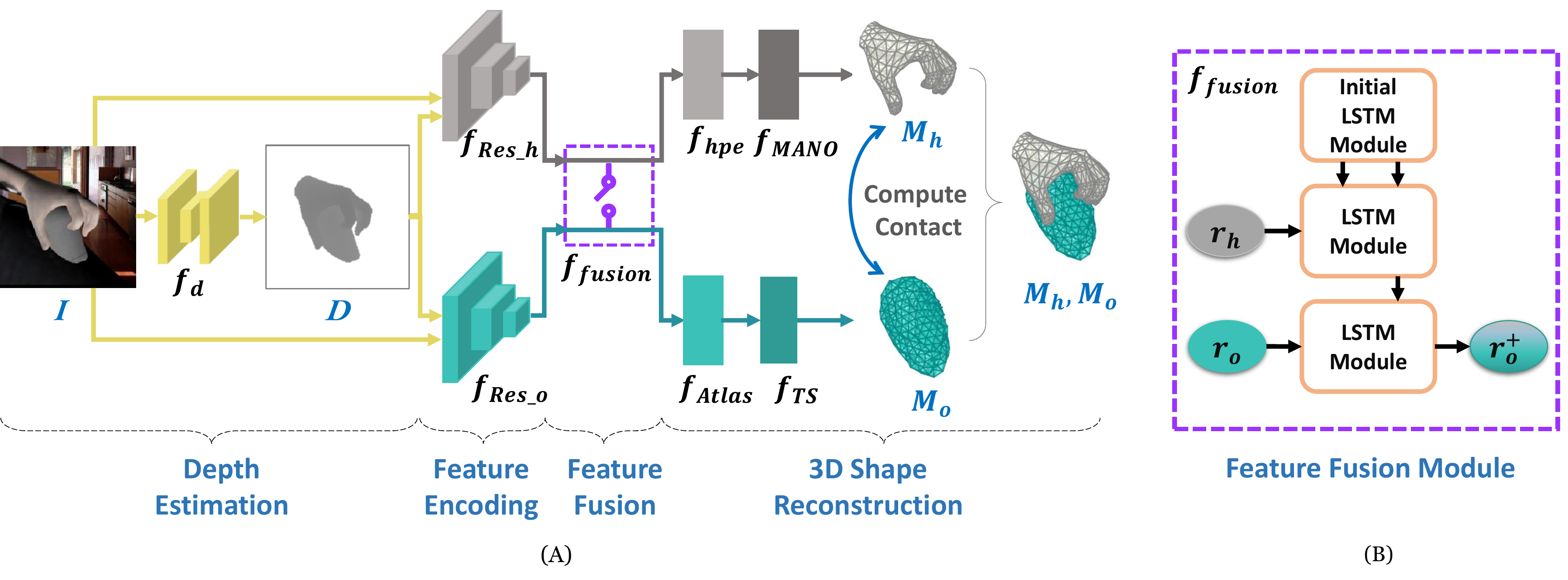}
	\vspace{-6mm}
	\caption{(A): The architecture of our proposed network.
	First, a depth map $D$ is estimated from the input image $I$ through the depth estimation module $f_d$.
	Then the depth map $D$, concatenated with the input image $I$, is fed into the rest network.
    Two separate encoders, $f_{Res\_h}$ and $f_{Res\_o}$, are used to extract the hand and object feature $\bm{r_{h}}$ and  $\bm{r_{o}}$, respectively.
    A feature fusion module $f_{fusion}$ is then employed to perform cross-branch communication.
	Finally, the fused features are used to reconstruct 3D meshes $M_h$ and $M_o$.
	(B): A variant of the feature fusion module $f_{fusion}$:
	The hand feature $\bm{r_{h}}$ and the object feature $\bm{r_{o}}$ are sequentially fed into LSTM modules to obtain the fused hand-aware object feature $\bm{r_o^+}$.
	Several other variants are presented in Fig. \ref{fig:3}.}
    \label{fig:2}
	}}
\end{figure*}

To capture and understand hand-object interaction, there are two commonly used methods to generate 3D shapes from 2D image inputs. One is to restore 3D hand pose and 6D object pose~\cite{brachmann2014learning}, and the 3D relationship can be inferred from sparse hand keypoints and object bounding boxes \cite{oberweger2019generalized, doosti2020hope}.
Another way is to directly reconstruct the pose and shape of hand and object, so that more detailed information can be obtained, such as the contact between the surfaces.
In this case, the 3D model of human hands can be restored by estimating hand pose and using a shape prior model \cite{taylor2014user,khamis2015learning,romero2017embodied}. And if a 3D model of the object is available, we can retrieve the object shape from object database and align it with the predicted 6D pose into camera coordinate system \cite{kokic2019learning,hasson20_handobjectconsist}.

The goal of our work is to recover high-quality 3D hand and object meshes from a single hand-object image without knowing object categories.
This task has been studied previously \cite{hasson2019learning,kokic2019learning}, but the state-of-the-art performance is still far from satisfactory due to the following challenges.
First, recovering 3D shape from a single-view image is an ill-posed problem, and we need to infer the missing depth information from 2D observations.
Second, the severe mutual occlusions between hands and objects coupled with  inter-finger occlusions make the reconstruction even more difficult.
Third, physical constraints need to be considered to avoid inferring infeasible interactions.
For example, surface penetration is usually not allowed and physical contacts are required when a hand holds an object.
However, it is challenging for the network to discover these constraints from the training data without providing domain knowledge.

Different from existing methods \cite{hasson2019learning,kokic2019learning} that reconstruct hand shape and object shape separately using state-of-the-art networks from these two individual tasks, we jointly estimate their 3D shapes and explore the reciprocity of two reconstruction tasks. Specifically, as shown in Fig.~\ref{fig:1}, our method first augments the RGB image into RGB-D representation by estimating the corresponding depth map and then feeds the RGB-D image into the feature encoding branches for hand and object individually. Inspired by the ``Block LSTM" used for sequential blocks connection in \cite{tian2019learning}, we use the connection information of structural components to benefit the joint reconstruction task. We modify the parallel joint reconstruction task by adding a step-by-step feature fusion process that propagates features in the latent space through a long short-term memory (LSTM) \cite{hochreiter1997long} block. Generally, for different scenarios and samples, the shape of hands conforms to some priors while the shape of objects can be more varied. In our setup, we found the hand part is relatively more robust than the object part because the hand part employs a low-dimensional parametric model learned from more than two thousand 3D scans~\cite{romero2017embodied}, while the object part uses less constrained vertex positions. Therefore, we use the more robust hand feature to enhance the object feature through an LSTM feature fusion module. Specifically, the hand feature and the object feature enter the LSTM module sequentially (see Fig.\ref{fig:2}-B) so that the later entered object feature is modulated by the LSTM state, which stores information from the hand feature. We further investigate several alternative feature fusion strategies in Section~\ref{section:as} and demonstrate the effectiveness of the LSTM-based module.
Finally, the hand feature and the enhanced object feature are decoded to regress 3D shapes of hand and object.

Our main contributions are summarized as follows:

$\bullet$ \hangindent 3em \, We design an end-to-end cross-branch network that significantly improves accuracy over existing methods for reconstructing 3D shapes of hand and the interacting object from a single RGB image.

$\bullet$ \hangindent 3em \, We propose the first approach to extensively investigate the feature fusion scheme for jointly reconstructing hands and objects.

$\bullet$ \hangindent 3em \, We introduce a depth estimation module to augment the RGB input into an RGB-D image, which effectively brings additional performance gains.

\section{Related Work}
This work is primarily related to the 3D reconstruction of hands and objects from hand-object interaction images.
Secondly, this work is relevant to hand pose and shape estimation.
Thirdly, it is also related to single view 3D object reconstruction. %
Moreover, the proposed LSTM feature fusion scheme is related to approaches that employ recurrent neural network (RNN) in 3D vision tasks.
We briefly review the related work in this section.\\
\textbf{3D parsing of hand-object image.}
Analyzing hand-object interaction in 3D is a challenging topic in computer vision, including estimating the 3D shape of hands in interaction with object \cite{romero2010hands,hamer2010object}, tracking hands while it is interaction with an object \cite{hamer2009tracking,ballan2012motion, tzionas2016capturing, sridhar2016real}, and reconstructing objects for in-hand scanning systems \cite{tzionas20153d}, etc.
Recently, the development of deep learning has further advanced the progress of this topic.
Recent learning-based works conduct joint hand-object pose estimation either from depth image~\cite{oberweger2019generalized} or from RGB image \cite{gao2019variational,hampali2019ho,tekin2019h+}, however, sparse joint estimation is probably not sufficient for reasoning about the hand model or hand-object contact.
Very recently, some works \cite{hasson2019learning,kokic2019learning,tsoli2018joint,hasson20_handobjectconsist,karunratanakul2020grasping} take into account 3D reconstruction of hand-object interaction.
Instead of tracking hand that interacts with a deformable object as in \cite{tsoli2018joint}, we focus on estimating 3D shapes of a hand and a rigid object.
Although \cite{kokic2019learning,hasson20_handobjectconsist} also model 3D hand-object shapes, the object categories are known in their settings.
In order to adapt to more types of objects, we directly estimate object shapes and do not assume that the model of the object is pre-defined.
Our proposed method is most similar to \cite{hasson2019learning} which uses a two-branch network to estimate 3D representation of hand and object from the RGB image.
There are two main differences between our method and \cite{hasson2019learning}.
First, we introduce a depth estimation module to estimate the depth map containing abundant structure information and use the RGB-D information for the 3D reconstruction.
Second, the hand-object relationship has not been well explored in \cite{hasson2019learning}, while we introduce the connection between the two branches through the proposed LSTM feature fusion module.\\
\textbf{Hand pose and shape estimation.}
Hand pose estimation has developed rapidly since the advance of commodity \mbox{RGB-D} sensors \cite{keselman2017intel,zhang2012microsoft}.
Significant progress in 3D hand pose estimation from either RGB image \cite{mueller2018ganerated,spurr2018cross,panteleris2018using,cai2018weakly,cai2019exploiting,iqbal2018hand,liang2013model,chen2021self} or depth map \cite{yuan2018depth,chen2019so,ge2018point,guo2017region,moon2018v2v,sun2015cascaded,tang2014latent} has been witnessed.
To better display the surface information of the hand, many works focus on producing a dense hand mesh which is achieved through depth input \cite{malik2018deephps} or RGB input \cite{ge20193d,hasson2019learning,zhang2019end,zimmermann2019freihand,boukhayma20193d,baek2019pushing,moon2020i2l}.
In our work, to better infer hand interactions with objects, we also focus on predicting accurate hand meshes from RGB image.
The MANO hand model~\cite{romero2017embodied}, whose parameters are regressed from a neural network, is employed to produce the hand joints and~meshes.\\
\textbf{Single-image 3D object reconstruction.}
Recovering 3D object shape from a single image is a challenging problem.
Recent approaches have attempted to represent objects in voxels \cite{choy20163d,wu2018learning}, point clouds \cite{fan2017point,lin2018learning}, octrees \cite{riegler2017octnet,tatarchenko2017octree}, or polygon meshes \cite{deprelle2019learning,gkioxari2019mesh,groueix2018atlasnet,kato2018neural,smith2019geometrics,wang2018pixel2mesh}.
Our work closely relates to methods that represent objects by meshes.
They usually generate 3D structures by deforming a set of primitive square faces \cite{groueix2018atlasnet} or a generic pre-defined mesh \cite{kato2018neural,wang2018pixel2mesh}.
Similar to AtlasNet~\cite{groueix2018atlasnet}, we also form 3D meshes by deforming a set of primitive square faces. A view-centered variant of AtlasNet is adopted to generate 3D objects from the hand-object images~\cite{hasson2019learning}.\\
\textbf{Recurrent neural network in 3D vision tasks.} RNN is well-accepted for its effectiveness in processing temporal sequence \cite{li2015constructing, sutskever2014sequence}. Recently, RNN becomes a powerful regulator in 3D vision tasks, such as human pose estimation \cite{lee2018propagating,liu2019feature,shahroudy2016ntu} and 3D object reconstruction \cite{zou20173d}.
Very recently, \cite{liu2019feature} presents a ConvLSTM to allow the propagation of contextual information among different body parts for 3D pose estimation.
\cite{tian2019learning} proposes 3D shape programs, which take a 3D shape as input and outputs a sequence of primitive programs, to describe the corresponding 3D shape of an object.
Unlike them, our goal is not to parse an individual structure by using the contextual information among structural parts.
Instead, we aim to link the information of two different individuals with an LSTM feature fusion module in our joint hand-object reconstruction task.

\section{Proposed Approach}
The goal of our method is to reconstruct both hand and object in mesh representation from an image where a hand is interacting with an object.
We propose a two-branch branch-cooperating network to solve the problem.
A depth map is predicted from the RGB image first, and then the RGB-D image, which is a concatenation of the original RGB image and the estimated depth map, is fed into the following reconstruction branches.
A two-branch network is adopted to estimate hand and object shapes, and a feature fusion module bridging the two branches is proposed to enhance the individually learned hand and object features (see Fig.~\ref{fig:2}).
In the following, we describe our proposed method in detail.

\subsection{Depth Estimation}
\label{sec:2.5d_inference}
Previous works~\cite{wu2017marrnet,zhang2018learning} show that utilizing depth map can bring benefits to the monocular RGB-based shape estimation task.
However, ground truth depth maps are not always available with RGB images. In this work, we employ a depth estimation module to augment the RGB input to an \mbox{RGB-D~input}.

A depth estimation module $f_{d}$ is used to estimate depth map $D$ from the color image $I$: $D = f_{d}(I)$.
Similar to \cite{wu2017marrnet}, ResNet-18 \cite{he2016deep} is used to encode the input image into latent representation, and a decoder is used to predict the depth map.
The decoder consists of four sets of $5 \times 5$ convolutional layers and four sets of $1 \times 1$ convolutional layers, and a ReLU layer is added after each convolutional layer.
The output is a one-channel depth map of the same size as the input image.+
We concatenate the estimated depth map $D$ with the source RGB image~$I$, and fed the RGB-D image into the following network.

The loss for this depth estimation task consists of two parts: the least absolute deviations loss (also referred as L1 distance) $\mathcal{L}_{1}$ and the structural similarity (SSIM) loss $\mathcal{L}_{ssim}$ (Eq.~\ref{eq:2}).
$\lambda_{ssim}$ is a weighting factor which is set to 1000 empirically.
We use $\frac{1}{SSIM}$ as a loss term to encourage the structural similarity between the predicted depth map and its ground truth \cite{wang2004image}.
The depth loss $\mathcal{L}_{D}$ is the summation of the least absolute deviation loss $\mathcal{L}_{1}$ and the structural similarity loss $\mathcal{L}_{ssim}$ (Eq.~\ref{eq:3}).
\begin{equation}
\label{eq:2}
    \mathcal{L}_{ssim} = \frac{1}{\lambda_{ssim} \cdot SSIM}
\end{equation}
\begin{equation}
\label{eq:3}
    \mathcal{L}_{D} = \mathcal{L}_{1} + \mathcal{L}_{ssim}
\end{equation}

Note that the depth values of the background are missing in the synthetic images, and therefore we only calculate the loss for the foreground regions during network training.

\subsection{Feature Encoding}
\label{sec:feature_en_bo}
The raw RGB image is first concatenated with the estimated depth map, and the four-channel RGB-D image is fed into the encoders of the two-branch network. Each branch takes the RGB-D image as input and extracts a latent representation by the ResNet-18 encoder~\cite{he2016deep} pre-trained on the ImageNet~\cite{russakovsky2015imagenet}. The hand and object representation is denoted as \mbox{$\bm{r_{h}} = f_{Res\_h}(I\oplus D)$} and $\bm{r_{o}} = f_{Res\_o}(I \oplus D)$ respectively, where $\oplus$ denotes concatenation.
\begin{figure*}[tb]
    \centering
    \makebox[0pt][c]{\parbox{1\textwidth}{%
    \centering
	\vspace{-6mm}
		\includegraphics[width=0.84\linewidth]{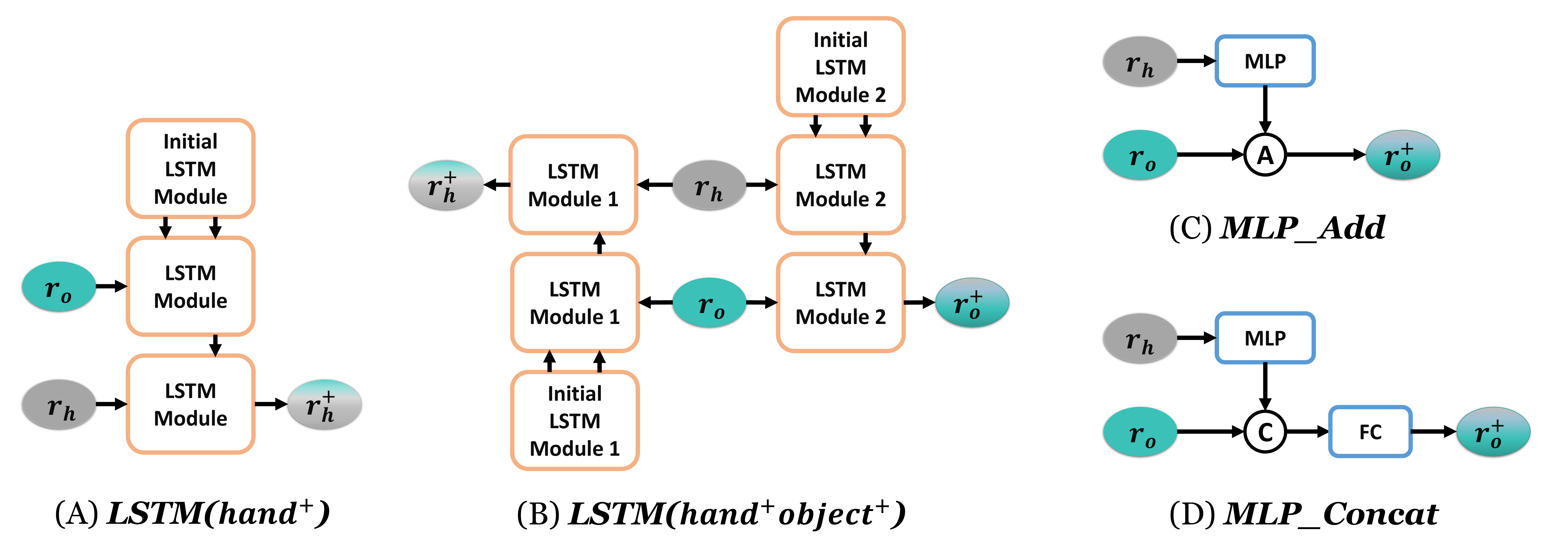}
    	\vspace{-4mm}
    	\caption{Alternative feature fusion architectures.
    	(A) \textbf{\textit{LSTM(\bm{$hand^+$})}}.
    	(B) \textbf{\textit{LSTM}(\bm{$hand^+object^+$})}.
    	(C) \textbf{\textit{MLP\_Add}}. (D) \textbf{\textit{MLP\_Concat}}.
    	}
	\label{fig:3}
	}}
	\vspace{-4mm}
\end{figure*}

\subsection{3D Shape Reconstruction}
\label{sec:3d_shape}
\subsubsection{Hand pose and shape estimation}
We regress model parameters from the latent representation and then use the MANO model as a network layer to predict hand pose and shape as  \cite{hasson2019learning,zhang2019end}.
Specifically, we feed the hand feature~$\bm{r_{h}}$ to the hand parameter estimator $f_{hpe}$ to regress hand parameters, including pose $\theta$ and shape $\beta$:
$\theta, \beta = f_{hpe}(\bm{r_{h}})$.
Then, a differentiable MANO hand model $f_{MANO}$ is used to compute hand joints $J_{h}$ and hand mesh $M_{h}$ according to $\theta$ and $\beta$:
$J_{h}, M_{h} = f_{MANO}(\theta, \beta)$.
$\theta$ determines the joint angles and $\beta$ adjusts the shape of the hand (see \cite{romero2017embodied} for more details).

To supervise the learning of hand estimation, we adopt a common joint position loss $\mathcal{L}_{J}$.
Meanwhile, a vertex position loss $\mathcal{L}_{M}$ is used to supervise the training of hand mesh when the ground truth hand meshes are available.
Both losses are calculated with the L2 distance between the predicted value and the ground truth value.
The shape regularization term is defined as $\mathcal{L}_{\beta} = {\parallel \beta - \bar{\beta} \parallel}^{2}$ to encourage the estimated hand model shape $\beta$ to be close to the average shape $\bar{\beta} = \vec{0} \in \mathbb{R}^{10}$. The pose regularization term is defined as $\mathcal{L}_{\theta} = {\parallel \theta - \bar{\theta} \parallel}^{2}$ to penalize the unreasonable pose which is far from the average pose $\bar{\theta} = \vec{0} \in \mathbb{R}^{30}$.
In summary, the overall loss for hand part $\mathcal{L}_{Hand}$ is the weighted sum of $\mathcal{L}_{M}$, $\mathcal{L}_{J}$, $\mathcal{L}_{\beta}$ and $\mathcal{L}_{\theta}$ (Eq.~\ref{eq:10}), where $\mu_{J}$, $\mu_{\beta}$ and $\mu_{\theta}$ are the~weighting~factors.
\begin{equation}
\label{eq:10}
    \mathcal{L}_{Hand} = \mathcal{L}_{M} + \mu_{J}\mathcal{L}_{J}  + \mu_{\beta}\mathcal{L}_{\beta}  + \mu_{\theta}\mathcal{L}_{\theta}
\end{equation}
\subsubsection{Object shape reconstruction}
Our object reconstruction branch is similar to AtlasNet~\cite{groueix2018atlasnet}.
In AtlasNet, the latent shape representation and a set of 2D points sampled uniformly in the unit square are fed into the network, and a collection of parametric surface elements is generated.
These parametric surface elements infer a surface representation of the shape.
AtlasNet is designed to reconstruct object meshes in a canonical view, and \cite{hasson2019learning} has validated its effectiveness in view-centered coordinate in the hand-object task.
Unlike AtlasNet, the latent representation is extracted from the RGB-D input in our method.
Then the AtlasNet block $f_{Atlas}$ decodes the latent representation $\bm{r_o}$ to an object mesh $M_{co}$ in the object canonical coordinate system (Eq.~\ref{eq:11}).
A multi-layer perceptron (MLP) neural network $f_{TS}$ is used to predict the parameters of translation $T$ and scale $s$  (Eq.~\ref{eq:12}).
Then, according to $T$ and $s$, we translate the regressed mesh $M_{co}$ into the final object mesh
$M_{o} = s M_{co} + T$.
\begin{equation}
\label{eq:11}
    M_{co} = f_{Atlas}(\bm{r_o})
\end{equation}
\begin{equation}
\label{eq:12}
    T, s = f_{TS}(\bm{r_o})
\end{equation}
We use a symmetric Chamfer loss $\mathcal{L}_{CD}$ as~\cite{groueix2018atlasnet, hasson2019learning}, which measures the difference between the predicted 642 vertices of the regressed object mesh $M_{o}$ and 600 points which is uniformly sampled on the surface of ground truth object mesh $M_{g}$~(Eq.~\ref{eq:CD}).
Except for comparing the object meshes $M_{o}$ with $M_{g}$ in the hand-relative coordinate, we also compute the Chamfer loss $\mathcal{L}_{CDcan}$ in the object canonical coordinate~(Eq.~\ref{eq:CDcan}), where $M_{co}$ is the regressed object mesh in Eq.~\ref{eq:11} and $M_{cg}$ is normalized from the ground truth mesh $M_{g}$ (Eq.~\ref{eq:gttrans}) by the ground truth object centroid $\hat{T}$ and the maximum radius~$\hat{s}$.
\begin{small}
\begin{equation}
\label{eq:CD}
    \mathcal{L}_{CD} = \frac{1}{2}(\frac{1}{600}\sum_{p\in M_{g}}\mathop{min}\limits_{q\in M_{o}}{\parallel p-q \parallel}^{2} + \frac{1}{642}\sum_{q\in M_{o}}\mathop{min}\limits_{p\in M_{g}}{\parallel q-p \parallel}^{2})
\end{equation}
\end{small}
\begin{small}
\begin{equation}
\label{eq:CDcan}
\begin{aligned}
    \mathcal{L}_{CDcan} = \frac{1}{2}(\frac{1}{600}\sum_{p\in M_{cg}}\mathop{min}\limits_{q\in M_{co}}{\parallel p-q \parallel}^{2} +\\
    \frac{1}{642}\sum_{q\in M_{co}}\mathop{min}\limits_{p\in M_{cg}}{\parallel q-p \parallel}^{2}){}
\end{aligned}
\end{equation}
\end{small}
\begin{equation}
\label{eq:gttrans}
    M_{cg} = (M_{g} - \hat{T})/\hat{s}
\end{equation}

The translation loss is defined as \mbox{$\mathcal{L}_{T} = {\parallel T - \hat{T} \parallel}^{2}_2$} and the scale loss is defined as \mbox{$\mathcal{L}_{s} = {\parallel s - \hat{s} \parallel}^{2}_2$}, where the ground truth object centroid $\hat{T}$ and the maximum radius $\hat{s}$ are computed in the hand-relative coordinates.
An edge-regularization loss $\mathcal{L}_{E}$ is used to penalize edges' length difference and a curvature-regularizing loss $\mathcal{L}_{L}$ is used to encourage reasonably smooth surface.
The overall loss for object part $\mathcal{L}_{Object}$ is the weighted sum of $\mathcal{L}_{CD}$, $\mathcal{L}_{CDcan}$, $\mathcal{L}_{T}$, $\mathcal{L}_{s}$, $\mathcal{L}_{L}$ and $\mathcal{L}_{E}$ (Eq.~\ref{eq:objloss}), where $\mu_{T}$, $\mu_{s}$, $\mu_{L}$, $\mu_{E}$ are the weighting factors.

\begin{equation}
    \mathcal{L}_{Object} = \mathcal{L}_{CD} + \mathcal{L}_{CDcan} + \mu_{T}\mathcal{L}_{T} + \mu_{s}\mathcal{L}_{s} + \mu_{L}\mathcal{L}_{L} + \mu_{E}\mathcal{L}_{E}
    \label{eq:objloss}
\end{equation}

\subsubsection{Connecting two branches via feature fusion module}
For this hand-object reconstruction task, we propose to link the independent features of the hand and the object branches.
To facilitate joint reconstruction, we propose to connect the two branches in latent space via a feature fusion module.
In our model, the hand representation $\bm{r_h}$ is more robust than the object representation $\bm{r_{o}}$ because the hand part uses the MANO model \cite{romero2017embodied}, which is a low-dimensional parametric model trained on the hand data from 2018 scans, thus it contains rich prior knowledge about the valid hand space.
In contrast, objects are represented using vertex positions optimized only by training data and therefore contain less shape prior information.
Therefore, we choose the hand feature $\bm{r_h}$ as the provider to enhance the object feature $\bm{r_{o}}$ by enabling the object branch to perceive information from the hand branch.
As shown in Fig.~\ref{fig:2}-B, we employ a two-timestep LSTM $f_{fusion}$ as the feature fusion module. The hand feature $\bm{r_{h}}$ is fed to the LSTM module at the first timestep, and then the state of the LSTM layer, storing the information of the current hand's pose and shape, is propagated into the object branch, resulting in enhanced hand-aware object feature $\bm{r_o^+} = f_{fusion}(\bm{r_{h}},\bm{r_{o}})$, where $\bm{r_{o}}$ is the original object feature. The two-timestep LSTM uses one recurrent layer and the number of features in the hidden state is 1000. After adding the feature fusion module, the object representation $\bm{r_o}$ is replaced by the enhanced object feature $\bm{r_o^+}$ in Eq.~\ref{eq:11} and Eq.~\ref{eq:12}. To better explore the effectiveness of the feature fusion module, we investigate several fusion strategies (Fig.~\ref{fig:3}) and compare their performance in~Section~\ref{section:as}.

Additionally, we take into account the contact between the hand and the object when recovering their meshes.
Different from 3D shape reconstruction from a hand-only or object-only image, jointly reconstructing the hand and the object needs to deal with the contact problem.
\cite{hasson2019learning} formulates the contact constraint as a differentiable loss $\mathcal{L}_{Contact}$, which consists of an attraction term and a repulsion term.
We adopt the same contact loss, and more details can be found in \cite{hasson2019learning}.
The overall loss function $\mathcal{L}$ of our proposed network is the weighted sum of the above mentioned four parts (Eq.~\ref{eq:allloss}), where the $\mu_{H}$, $\mu_{O}$, $\mu_{C}$ are the~weighting~factors.
\begin{equation}
\label{eq:allloss}
    \mathcal{L} = \mathcal{L}_{D} + \mu_{H}\mathcal{L}_{Hand} + \mu_{O}\mathcal{L}_{Object} + \mu_{C}\mathcal{L}_{Contact}
\end{equation}

\section{Experiments}
In this section, we first present datasets, evaluation metrics (Section~\ref{sec:dataset}), and implementation details (Section~\ref{section:id}) of our experiments.
Then, we analyze the effect of the proposed modules (Section~\ref{section:as}) and evaluate the overall performance of our method (Section~\ref{section:results}).
\subsection{Datasets and Evaluation Metrics}
\label{sec:dataset}
We evaluate our method on three publicly available datasets: a synthetic third-person perspective dataset, a real-world egocentric dataset, and a real-world third-person perspective dataset.\\
\textbf{ObMan Dataset:}
The ObMan dataset \cite{hasson2019learning} is a large-scale synthetic dataset containing hand-object images, in which the hand is generated through the MANO model~\cite{romero2017embodied} and the object is sampled from the ShapeNet dataset \cite{chang2015shapenet}.
More than 2000 object meshes are presented in the ObMan dataset whose objects are selected from eight categories in the ShapeNet dataset.
The plausible grasps are generated by using the GraspIt software \cite{miller2004graspit}.
The ObMan dataset contains 141K training frames and 6K test frames. For each frame, it provides the RGB-D image, 3D hand and object meshes, and 3D hand keypoints.
\\
\textbf{First-Person Hand Action Benchmark Dataset (FHB):}
The FHB dataset \cite{garcia2018first} collects RGB-D video sequences of daily hand action categories with hand-object interaction.
3D hand joints and 6D object poses are automatically annotated.
Following \cite{hasson2019learning}, a subset of FHB named FHBc containing around 5K frames for training and 5.6K frames for testing is used in the experiments.
The same object categories appear in both the training and~test~set.\\
\textbf{Hands in Action Dataset (HIC):} Four sequences in the HIC dataset \cite{tzionas2016capturing} are used, where a hand interacts with an object.
Similar to \cite{hasson2019learning}, we only use frames in which the hand is less than 5mm away from the object. We utilize sequence 15 and sequence 20 as the training set, and sequence 19 and sequence 21 for evaluation. The hand and object meshes are provided.

To evaluate the hand estimation results, we use three metrics:
(i) MPJPE: the mean per joint position error (MPJPE) in the Euclidean space for all joints on all test frames (in $mm$);
(ii)~HME: the hand mesh error (HME) is the average error in the Euclidean space between the corresponding vertices of the hand mesh on all test frames (in $mm$);
(iii) 3D PCK: the percentage of correct key points (PCK) whose joint error does not exceed a certain threshold.

To evaluate the object reconstruction results, we report the surface-to-surface distance between the predicted mesh and the ground truth mesh.
Specifically, the symmetric Chamfer error (CD) (Eq.~\ref{eq:CD}) is calculated on 600 points sampled on the ground truth mesh and all 642 vertices of the estimated mesh \cite{hasson2019learning,groueix2018atlasnet} (in $mm\times mm$).

To evaluate function of the contact between the hand and the object, we report the penetration depth (PD) which is the maximum penetration (in $mm$) between the hand and the object averaged on all the test frames as \cite{hasson2019learning}.
The maximum penetration is defined as the maximum distance (in $mm$) between the surfaces of the two meshes in the intersection space if exist. Otherwise, the error is 0 if the two meshes do not penetrate each other.

\subsection{Implementation Details}
\label{section:id}
We use Adam \cite{kingma2014adam} to optimize the network with batch size~16. The proposed network consists of three parts, i.e., the depth estimation part, the hand branch, and the object branch (together with the feature fusion module). Instead of directly optimizing the whole network, we adopt a stage-wise training strategy to train the network.

To train our model on the ObMan dataset, we divide the training procedure into several stages. We first train the depth estimation part for 90 epochs, we initialize the learning rate to $3\times10^{-4}$ and decrease it to $10^{-4}$ and $10^{-5}$ after every 30 epochs. Then the depth estimation part is frozen. The hand branch is trained for 250 epochs with the learning rate initialized to $10^{-4}$ and decreased to $10^{-5}$ at epoch 200. After that, the hand branch is frozen and the object branch is trained in the same procedure as the hand branch. Finally, we fine-tune the hand and object branches with the contact loss for 50 epochs by setting the learning rate to $10^{-5}$.

For the small FHB$c$ dataset, the same training procedure is used as that on the ObMan dataset except that the hand and object branches are initialized with the weights pre-trained on the ObMan dataset.
The two-branch 3D reconstruction part is trained for 150 epochs with learning rate deceased at epoch 100.
The weight of the depth map estimation part is randomly initialized since the regular depth map is used in the FHB{c} dataset while the depth foreground map is used in the ObMan dataset.

For the HIC dataset, the whole network is initialized with the weights pre-trained on the ObMan dataset. And we fine-tune the network for 250 epochs with the learning rate initialized to $10^{-4}$ and decreased to $10^{-5}$ at epoch 200. To supervise the training, we jointly use the hand loss $\mathcal{L}_{Hand}$, object loss $\mathcal{L}_{Object}$, and contact constraint $\mathcal{L}_{Contact}$.

For the hand estimation Eq.~\ref{eq:10}, the weighting factors are set as $\mu_J=0.167$, $\mu_{\beta}=0.167$ and $\mu_{\theta}=0.167$. For the object shape reconstruction  Eq.~\ref{eq:objloss},
$\mu_T=0.167$, $\mu_s=0.167$, $\mu_L=0.1$ and $\mu_E=2$. For the overall loss function Eq.~\ref{eq:allloss}, we set $\mu_H=0.001$, $\mu_O=0.001$ and $\mu_C=0.1$.

\subsection{Ablation Study}
\label{section:as}
We present an ablative analysis to evaluate the effectiveness of the proposed modules.
We analyze the effects of depth estimation, feature fusion module, two-branch pipeline, and contact computation. Table~\ref{table:ab} shows the results of different versions of our model.

\subsubsection{\textbf{Effect of depth estimation module}}
\begin{figure}[tb]
\centering
\vspace{-4mm}
		\centering
		\includegraphics[width=0.68\linewidth]{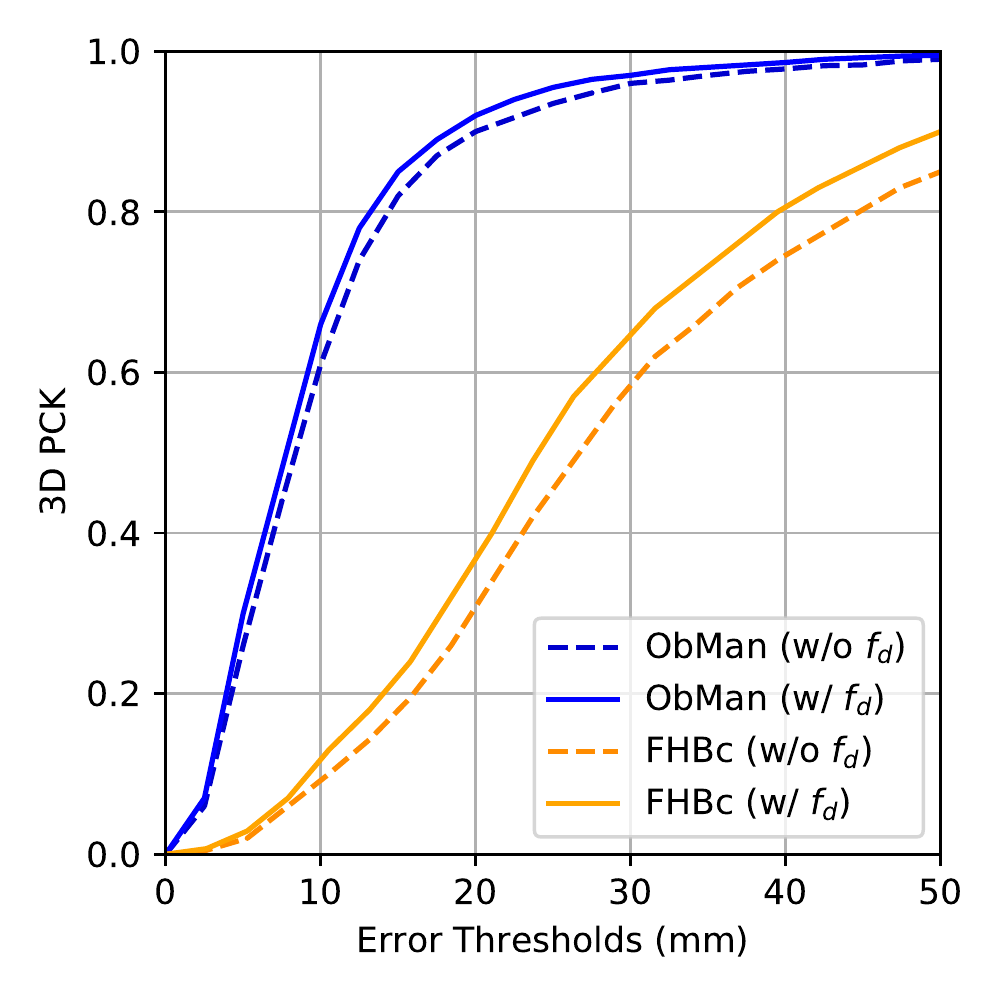}
	\vspace{-4mm}
	\caption{Comparison of whether to use the depth estimation module on the FHB$c$ dataset and the ObMan dataset.
	The proportion of correct keypoints over different error thresholds are presented.}
	\label{fig:de}
	\vspace{-2mm}
\end{figure}

To illustrate the effect of depth estimation, we conduct a comparison on both datasets to evaluate the hand estimation task.
For the setting without the depth estimation, we use an empty map to replace the estimated depth map and leave other components unchanged.
We evaluate the performance by the percentage of correct keypoints (PCK).
Note that PCK evaluates all joints of the test set, and here we follow \cite{zimmermann2019freihand} and use the error thresholds from 0 to 50mm in Fig.~\ref{fig:de}.
As shown in Fig.~\ref{fig:de}, the performance is improved compared to the method without the depth estimation module.
From the curves, we can see that the PCK is improved by about 10\% compared to the method without the depth estimation module when the error threshold is between 20mm and 40mm on the FHBc dataset, and the PCK is improved by about 3\% when the error threshold is between 10mm and 30mm on the ObMan dataset.
Apart from evaluating the improvement on hand estimation, we also report the improvement on object reconstruction of the ObMan dataset. Comparing index 2 and index 5 in Table~\ref{table:ab}, after removing the depth estimation module $f_d$ from index 2, the object CD will increase by 4\%, the hand MPJPE will increase by 6.2\%, and the hand HME will increase by 6.9\%.
Intuitively, the depth map is complementary to RGB image as it provides abundant and more direct structural information, which is beneficial for the inference of 3D reconstruction.
It is worth mentioning that the introduced depth estimation module does not need additional ground-truth information during training, as its supervision can be derived from ground-truth 3D meshes.

\begin{figure}[tb]
		\includegraphics[width=\linewidth]{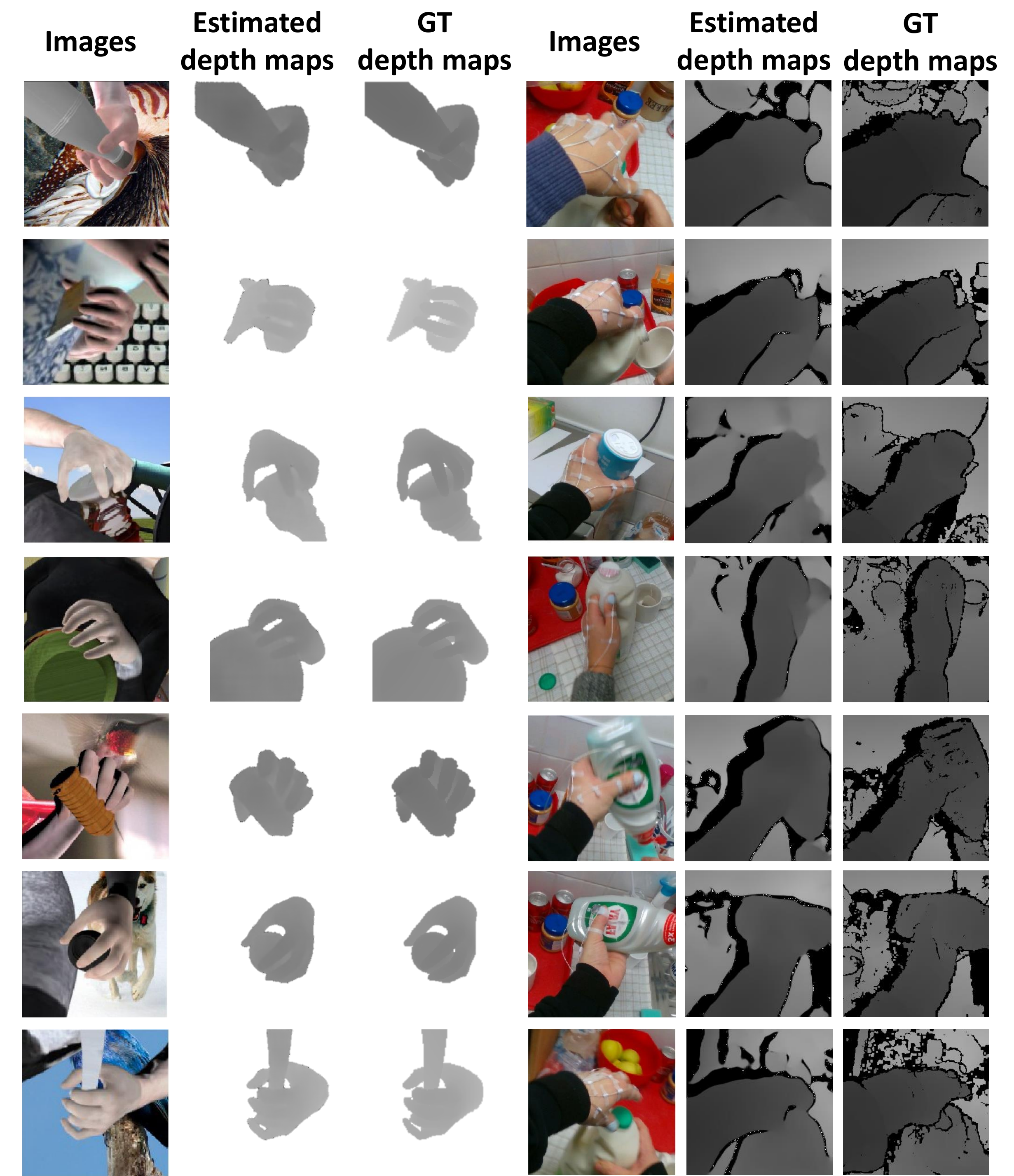}
	\vspace{-6mm}
	\caption{Qualitative visualization of the depth estimation on the ObMan dataset (left) and the FHB$c$ dataset (right).
	}
	\label{fig:depth}
	\vspace{-4mm}
\end{figure}

\subsubsection{\textbf{Comparison of different feature fusion modules}}
\label{subsub:different_fusion}
In this section, we conduct experiments to validate the effectiveness of the feature fusion module. Several alternative architectures of the module are investigated.
When no fusion is used, the hand branch and the object branch are trained separately (see \textbf{\textit{Baseline}} in Table~\ref{table:self-com}). Three fusion strategies are tested based on the provider and receiver of the cross-branch information.
(i)~\textbf{\textit{LSTM(\bm{$object^+$})}}:
LSTM means the proposed LSTM feature fusion module is used to fuse the features and "+" after object means the object feature is enhanced by fusing information from the other branch, which is the hand branch in this case (See Fig.~\ref{fig:2}-B).
(ii) \textbf{\textit{LSTM(\bm{$hand^+$})}}: The hand feature is enhanced by fusing information from the object branch via the LSTM feature fusion module (See Fig.~\ref{fig:3}-A);
(iii) \textbf{\textit{LSTM(\bm{$hand^+object^+$})}}: We use two LSTM modules, one to enhance object by hand and vice versa (See Fig.~\ref{fig:3}-B). Unlike the branch-wise training scheme in \textbf{\textit{LSTM(\bm{$hand^+$})}} and \textbf{\textit{LSTM(\bm{$object^+$})}}, the hand and object branches are trained together with mutual enhancement.
As shown in Table~\ref{table:self-com}, the object CD error of \textbf{\textit{LSTM(\bm{$object^+$})}} decreases by 5\% with the object feature enhanced by perceiving information from the hand branch, and the MPJPE of \textbf{\textit{LSTM(\bm{$hand^+$})}} decreases by 2\% with the hand feature enhanced. When two individual LSTM feature fusion modules are used to enhance both features (as \textbf{\textit{LSTM(\bm{$hand^+object^+$})}}, the results show that mutual fusion helps to improve the shape of the object but makes the hand worse.
We think that the more robust hand feature helps the object feature when they are fused and trained with each other.  Meanwhile, the object feature drags the hand feature down to a certain extent, resulting in slightly worse hand estimation and better object estimation. As for the choice of the direction of information transmission, we think it is a trade-off between improvements of the hand and the object. The \textbf{\textit{LSTM(\bm{$object^+$})}} is chosen in this paper because both hand and object get good results.

\begin{table*}[tb]
\centering
\makebox[0pt][c]{\parbox{1\textwidth}{
    \vspace{-0.25in}
    \begin{minipage}[t]{0.67\hsize}
    \caption{Evaluation results on the ObMan dataset in terms of the fusion strategy and whether to perform contact computation.
    {\rm ``\textit{Baseline}"} in this table indicates a separate two-branch network with the depth estimation module and without using the fusion module (defined in Section~\ref{subsub:different_fusion}).
    $\star$ indicates end-to-end training, and other settings are trained through a stage-wise approach (in Section~\ref{section:id}).}
    \label{table:self-com}
    \centering
    \begin{tabular}{lcc}
		\hline
		
		\hline
		Method & MPJPE $\downarrow$ & CD $\downarrow$\\
		\hline
		
		\hline
		\textit{Baseline} & 9.7 & 426.1\\
		\textit{Baseline+LSTM($hand^+$)} & 9.5 & 426.1\\
		\textit{Baseline+LSTM($object^+$)} & 9.7& 405.1 \\
		\textit{Baseline+LSTM($hand^+object^+$)} & 10.1 & 392.6 \\
		\hline
		\textit{Baseline+contact} & 9.7 & 422.5\\
		\textit{Baseline+LSTM($object^+$)+contact \textbf{(Ours)}} & 9.6 & 403.8\\
		\hline
		\textit{Baseline+LSTM($object^+$)+contact \textbf{(Ours)}}$\star$ & 10.1 & 391.5\\
		\hline
		
		\hline
	\end{tabular}
	
    \end{minipage}
    \hfill
    \begin{minipage}[t]{0.33\hsize}
    \caption{Object reconstruction results on the ObMan dataset with different feature fusion modules.
    }
    \label{table:3dobj}
        \centering
        \begin{tabular}{lc}
			\hline
			
			\hline
    		Method &  CD $\downarrow$\\
			\hline
			
			\hline
			\textit{Baseline} & 426.1\\
			\textit{Baseline+MLP\_Add($object^+$)} & 437.2
			\\
			\textit{Baseline+MLP\_Concat($object^+$)} & 428.1
			\\
			\textit{Baseline+LSTM($object^+$)} & 405.1
			\\
			\textit{Baseline+LSTM\_2layers($object^+$)} & 402.3
			\\
			\hline
			
			\hline
		\end{tabular}
    \end{minipage}
    }}
\end{table*}

\begin{table*}[tb]
\centering
\makebox[0pt][c]{\parbox{1\textwidth}{
        \begin{minipage}[t]{0.7\hsize}
        \caption{
        Ablation studies for different components used in our method on the ObMan dataset. {The depth estimation module $f_d$ contributed the most to the hand shape and the feature fusion module $f_{fusion}$ contributes the most to the object shape.}}
        \label{table:ab}
        \vspace{-2mm}
        \centering
    	\begin{tabular}{c|c|c|c|c|ccc}
    		\hline
    		
    		\hline
    		\centering
    	    \multirow{2}{*}{Index} & \multirow{2}{*}{$f_d$} & Separate & \multirow{2}{*}{$f_{fusion}$}& \multirow{2}{*}{Contact} & \multirow{2}{*}{MPJPE $\downarrow$}  & \multirow{2}{*}{HME $\downarrow$}  & \multirow{2}{*}{CD $\downarrow$} \\
    		~&~&Encoders&~&~&~&~&~\\
    		\hline
    		
    		\hline
    		1&$\checkmark$ &  $\checkmark$& $\checkmark$& $\checkmark$& \textbf{9.6}& \textbf{9.8} & \textbf{403.8}\\
    		2&  $\checkmark$&$\checkmark$ & $\checkmark$& $\times$&9.7 & 10.1 & 405.1\\
    		3&  $\checkmark$&$\checkmark$ & $\times$& $\times$& 9.7 & 10.0 &426.1\\
    		4&  $\checkmark$&$\times$ & $\times$& $\times$&10.5 & 10.8 &404.9 \\
    		5&  $\times$&$\checkmark$ & $\checkmark$& $\times$& 10.3 & 10.8 & 421.3\\
    		6&  $\times$ &$\checkmark$ & $\times$& $\times$&10.3 & 10.8 & 860.6 \\
    		\hline
    		
    		\hline
    	\end{tabular}
        \end{minipage}
        \hfill
        \begin{minipage}[t]{0.26\hsize}
        ~\\
        ~\\
        \caption{Comparison of MPJPE (mm) on the ObMan dataset and the FHBc dataset.}
        \label{table:hand}
        \begin{tabular}{ccc}
		\hline
		
		\hline
		\centering
		Method & FHBc \quad & ObMan \quad\\
		\hline
		
		\hline
		\cite{hasson2019learning}(w/o contact) &28.1&11.6\\
		\cite{hasson2019learning}(w/ contact) &28.8&11.6\\
		\hline
		Ours(w/o contact) &27.9&9.7\\
		Ours(w/ contact) &\textbf{27.5}&\textbf{9.6}\\
		\hline
		
		\hline
	\end{tabular}
    \end{minipage}
}}
\end{table*}

\begin{figure}[t]
    \centering
    \includegraphics[width=0.8\linewidth]{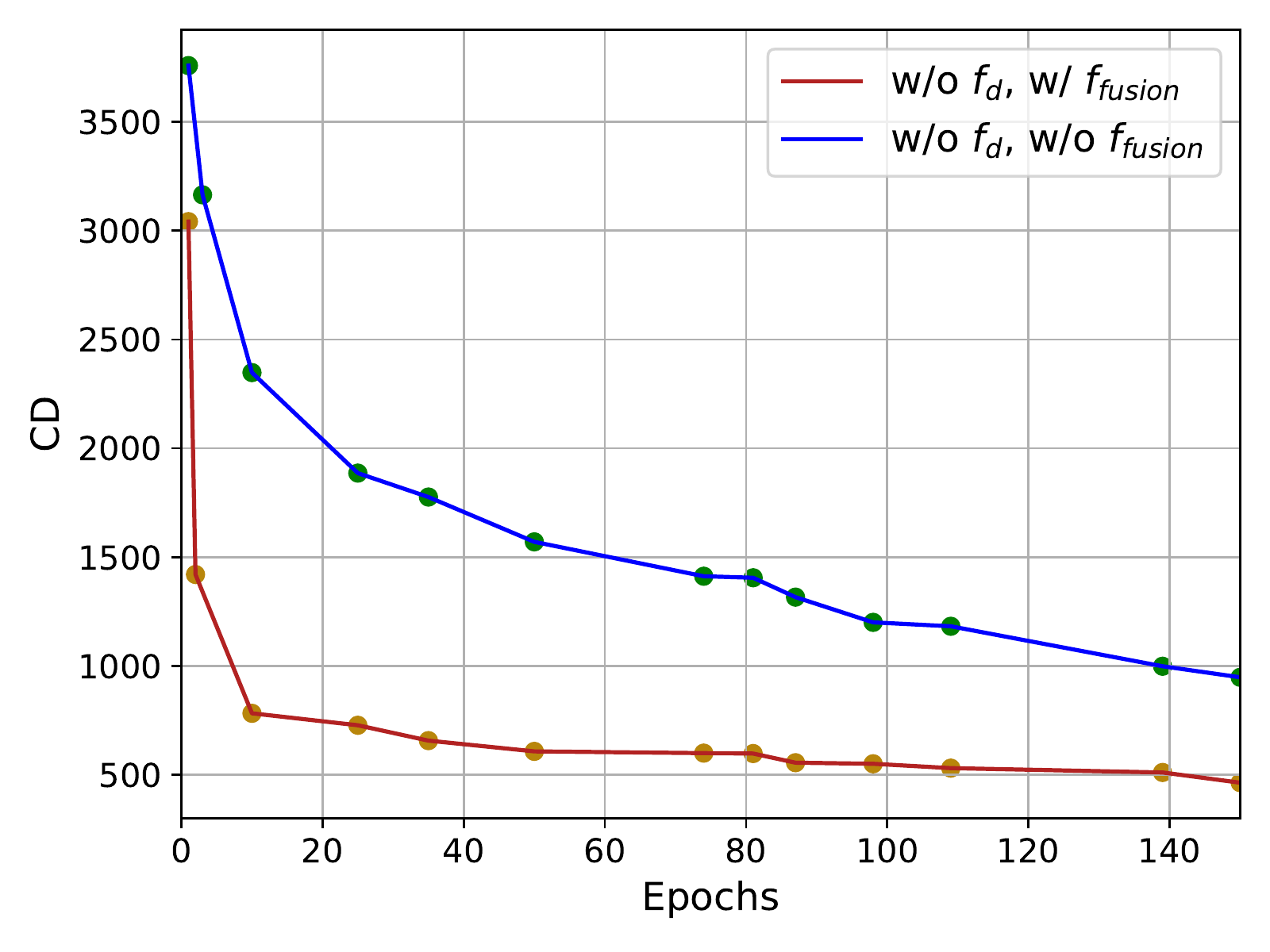}
    \vspace{-4mm}
    \caption{Ablation of the LSTM fusion module on the ObMan dataset. This plot shows the first 150 epochs of training: the $f_{fusion}$ makes the object CD error in the evaluation set decrease fast.
	}
	\label{fig:cd_curve}
    \vspace{-2mm}
\end{figure}

Apart from the proposed LSTM feature fusion module, we conduct ablative experiments to study the commonly-used MLP-based feature fusion modules, including concatenation and element-wise addition.
We also use a stacked LSTM in the LSTM module to see if more capacity can further improve the performance.
Three alternative feature fusion strategies are tested to enhance the object feature using information from the hand feature.
(i) \textbf{\textit{MLP\_Add(\bm{$object^+$})}}:
The object feature is enhanced via element-wise addition with information from the hand branch through a MLP mapping (See Fig.~\ref{fig:3}C);
(ii)~\textbf{\textit{MLP\_Concat(\bm{$object^+$})}}:
The object feature is enhanced via concatenation with information from the hand branch through an MLP mapping, and then the enhanced object feature is converted to a fixed size vector via a fully connected (FC) layer (See Fig.~\ref{fig:3}D);
(iii) \textbf{\textit{LSTM\_2layers(\bm{$object^+$})}}:
The same feature fusion method is used as in \textbf{\textit{LSTM(\bm{$object^+$})}}, while two LSTMs are stacked together to form a stacked LSTM in the LSTM module.
As shown in Table~\ref{table:3dobj}, the object reconstruction results become worse after using the MLP-based modules, while the \textbf{\textit{LSTM}} and \textbf{\textit{LSTM\_2layers}} can greatly improve the performance of the object reconstruction.
Note that in our experiments, MLP-based modules use similar parameters as in \textbf{\textit{LSTM(\bm{$object^+$})}}, while \textbf{\textit{LSTM\_2layers(\bm{$object^+$})}} uses double parameters.
Among these modules, the LSTM-based modules are most helpful and the LSTM module stacks two LSTMs is a little better than another but it has double parameters in the feature fusion module.
We use \textbf{\textit{LSTM(\bm{$object^+$})}} in~this~paper.

As shown in Fig~\ref{fig:cd_curve}, we compare the decreasing speed of the object CD in the evaluation set. To better evaluate the effect of the feature fusion module \textbf{\textit{LSTM(\bm{$object^+$})}}, the depth estimation module is removed. It can be seen from the curve that the proposed feature fusion module greatly accelerates the speed of the object CD error reduction.
Finally, if both the depth estimation module and the feature fusion module are removed from our model (shown as index 6 in Table~\ref{table:ab}), the reconstruction performances of both hands and objects will drop dramatically.

\subsubsection{\textbf{Effect of two-branch encoding}}
In addition to the above methods of encoding features with two encoders and then fusing cross-branch features, we also implemented an extreme bridging method, that is, the hand branch and the object branch shares the same features encoded by one encoder.
In this setup, a ResNet-34 is used as a shared encoder~$f_{Res\_s}$ in order to balance the total trainable parameters to make a fairer comparison.
As present in Table~\ref{table:ab} (index 4), we use a shared backbone and remove the hand encoder, object encoder and the fusion module from index 2. By replacing the separate encoders with the shared backbone network, comparable results can be obtained on object reconstruction, while the hand MPJPE increased by 10\%. We think that each branch specializes in its task while the fusion module helps with perceiving cross-branch information in the two-branch branch-cooperation pipeline, thus the two-branch pipeline is more efficient than the shared encoder setup for this joint hand-object reconstruction task.

\subsubsection{\textbf{Contact loss and its connection to feature fusion module}}
To study the effect of the feature fusion module and the contact term, different modules are employed in the experiments.
\textbf{\textit{Baseline+contact}}: We train the hand and object branches without feature fusion as the \textbf{\textit{Baseline}} and then add a contact loss to fine-tune both branches;
\textbf{\textit{Baseline+LSTM(\bm{$object^+$})+contact~(Ours)}}:
“+contact” means a contact loss is used to fine-tune both the branches.

Note that each stage has been trained to convergence.
As shown in Table~\ref{table:self-com}, comparing \textbf{\textit{Baseline+contact}}, \textbf{\textit{Baseline+LSTM(\bm{$object^+$})}} and \textbf{\textit{Baseline+ LSTM(\bm{$object^+$})+contact (Ours)}}, the object error decreases by 0.8\%, 4.9\% and 5.2\% accordingly.
The proposed LSTM module and the contact computing both are ``bridges" to connect hand and object, the results reveal that our proposed LSTM block is more helpful than the contact loss in reducing the object reconstruction error, and the combination of the LSTM module and the contact loss effectively improves performance.
Both the proposed LSTM feature fusion module and the contact constraint link two branches of hand and object, but they achieve this goal in different spaces.
The feature fusion module connects branches in the feature space while the contact constraint is implemented directly on the meshes.
We think that the connection in the feature space is easier to learn, so the proposed LSTM module contributes more than the contact loss.
In summary, the LSTM block can cooperate well with the contact term and helps to get better shapes from the hand-object image, and \textbf{\textit{Baseline+LSTM(\bm{$object^+$})+contact (Ours)}} achieves best overall performance on hand-object reconstruction task.
\subsubsection{\textbf{Comparison of training schemes}} As illustrated in Section~\ref{section:id}, we design a stage-wise training scheme to train the network components separately. Our network can also be trained in an end-to-end manner in which the whole network is trained for 250 epochs with the learning rate initialized to $10^{-4}$ and decreased to $10^{-5}$ at epoch 200. The final MPJPE and CD is 10.1 and 391.5, respectively (see Table~\ref{table:self-com}). Similar to the results of \textbf{\textit{Baseline+LSTM(\bm{$hand^+object^+$})}}, the end-to-end version gets better object reconstruction performance, but the hand performance is not as good as the stage-wise version. We think this may be because the poor object feature drags the hand feature down to a certain extent while the hand feature is worse than that in the fixed trained hand branch. Note that except for this set of end-to-end training results, other experiments all use a stage-wise training scheme.

\begin{figure}[tb]
\centering
		\centering
		\includegraphics[width=0.92\linewidth]{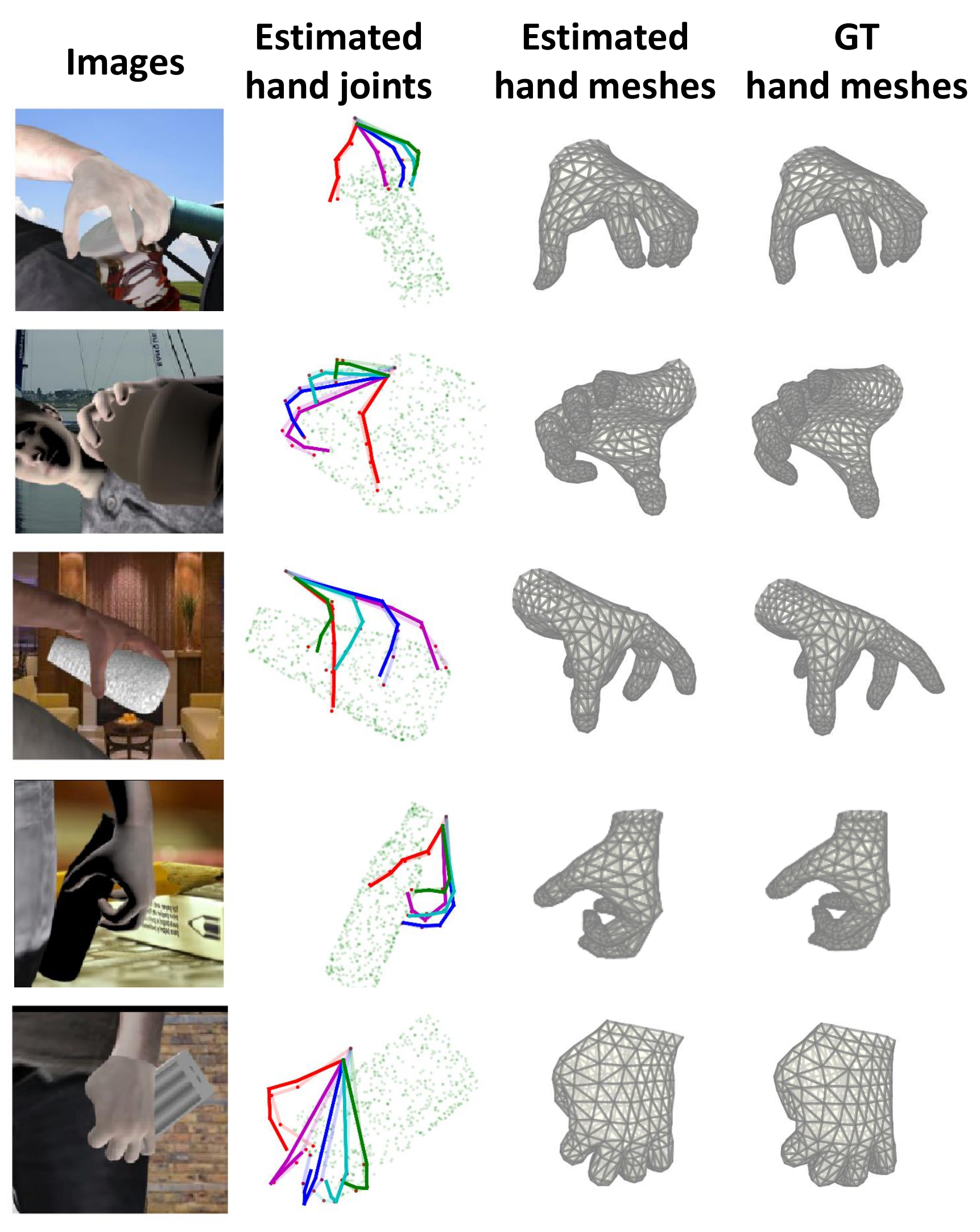}
	\vspace{-1mm}
	\caption{Qualitative comparison between our estimated hand meshes and ground truth meshes on the ObMan dataset. In the column of hand joints, we use the skeleton with opaque color to represent the estimated hand joints and the skeleton in semi-transparent color to represent the ground truth. Green points are randomly sampled from the ground truth object meshes.
	Results show that the estimated hand joints and meshes are close to the ground truth.
	Note that scales are changed in the visualization.
	}
	\label{fig:hand}
	\vspace{-8mm}
\end{figure}

\subsection{Results}
\label{section:results}
In this section, we first present results of depth estimation (\ref{subsec:depth}), hand recovery (\ref{subsec:hand}), object recovery (\ref{subsec:object}), and shape penetration (\ref{subsec:pene}) on the ObMan and FHBc datasets. Then we use the HIC dataset to evaluate the performance of fine-tuning a synthetic dataset pre-trained model to the small real-world dataset (\ref{subsec:finetune}). In addition, we present some visualization results.
\subsubsection{Performance of depth estimation}
\label{subsec:depth}
For depth estimation, we use a simple encoder-decoder structure to directly predict a depth image from the input monocular RGB input.
We estimate depth foregrounds on the ObMan synthetic dataset and produce regular depth maps on the FHB$c$ dataset.
The qualitative results are shown in Fig.~\ref{fig:depth}.
We find that the estimated depth images preserve intrinsic information about hand-object shape with smooth surfaces.
Note that when training the depth estimation part, the weights of the network are randomly initialized on both datasets.

\subsubsection{Performance of hand estimation}
\label{subsec:hand}
We compare the hand estimation result of our method with the state-of-the-art method \cite{hasson2019learning}.
The mean error of all joints over all test frames is presented in Table~\ref{table:hand}.
On the challenging FHB$c$ dataset which only provides a small amount of similar images, our method outperforms \cite{hasson2019learning} in both settings.
Our method obtains hand pose error of 27.9mm while the baseline method \cite{hasson2019learning} is 28.1mm under the setting without contact loss.
With the contact loss, our method obtains an error decreasing by 4.5\% (28.8mm vs 27.5mm), while the MPJPE of \cite{hasson2019learning} increased by 2.5\%.
On the large-scale ObMan dataset, the performance of our method is 16.4\% better than \cite{hasson2019learning}, and we obtain a MPJPE of 9.7mm while their MPJPE is 11.6mm.
It is worth noting that after adding the contact loss, the accuracy of hand joints remains unchanged or slightly decreases in \cite{hasson2019learning}, while it has an obvious improvement in our method.
We think this is because the proposed LSTM feature fusion module helps the network reconstructs shapes with more reasonable relative positions, and the shapes with better relative positions can better initialize the fine-tuning stage with the contact loss.
We present the visualized samples from the testing set in Fig.~\ref{fig:hand}.

\begin{figure*}[t]
\centering
\makebox[0pt][c]{\parbox{1\textwidth}{
	\vspace{-11mm}
	\centering
		\includegraphics[width=0.9\linewidth]{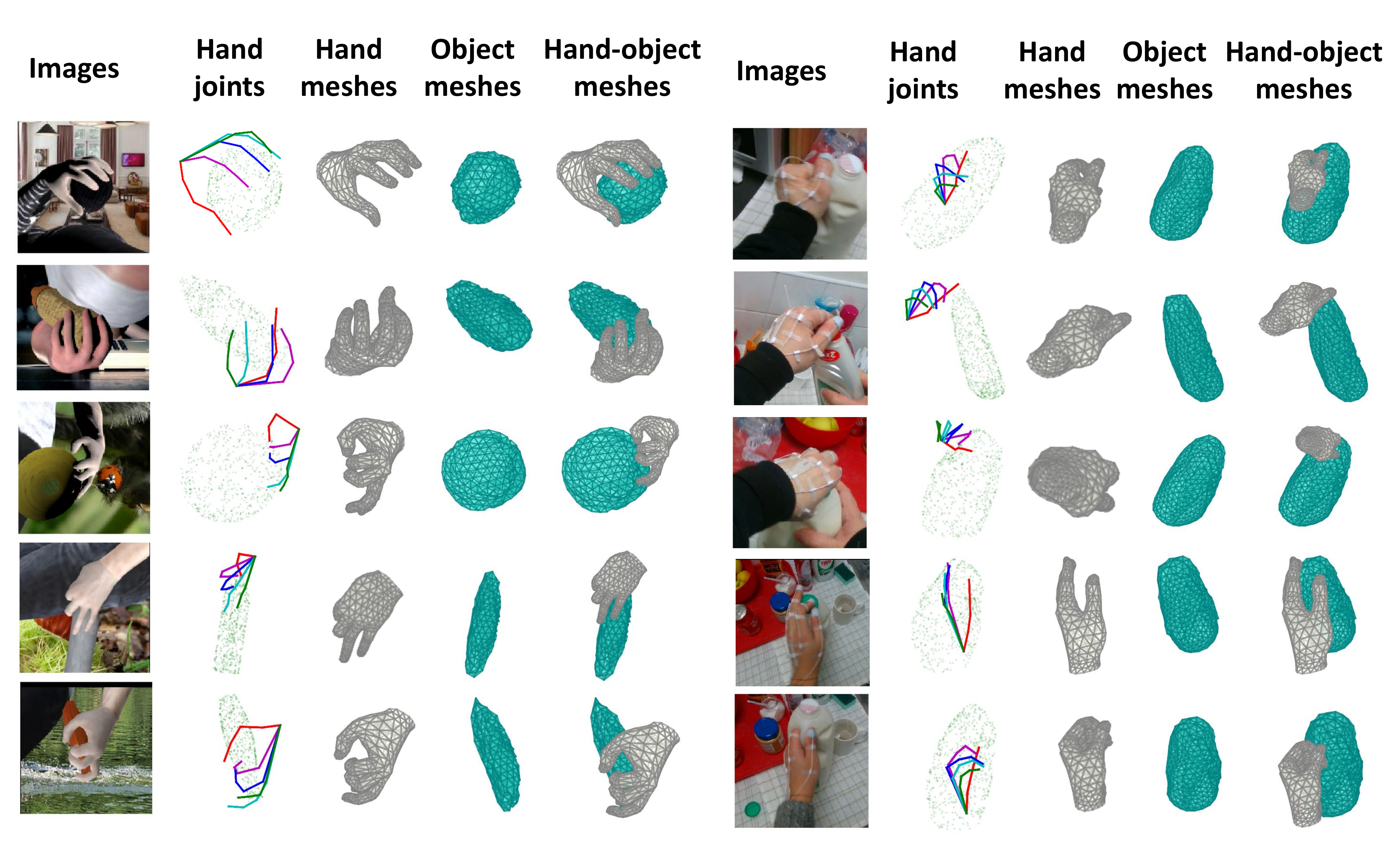}
	\vspace{-6mm}
	\caption{Qualitative comparison of our method on the ObMan dataset (left) and the FHB$c$ dataset (right).
	From left to right: input, estimated 3D hand pose, estimated hand mesh, estimated object mesh, estimated hand and object meshes.
	In the column of hand joints, we sample points from the ground truth object meshes to reveal the relationship between our estimated hand pose and the object.
	Note that we changed scales in the visualization. }
	\label{fig:hoshapes}
	\vspace{-0.13in}
}}
\end{figure*}

\begin{figure}
    \centering
    \vspace{-3mm}
    \includegraphics[width=1\linewidth]{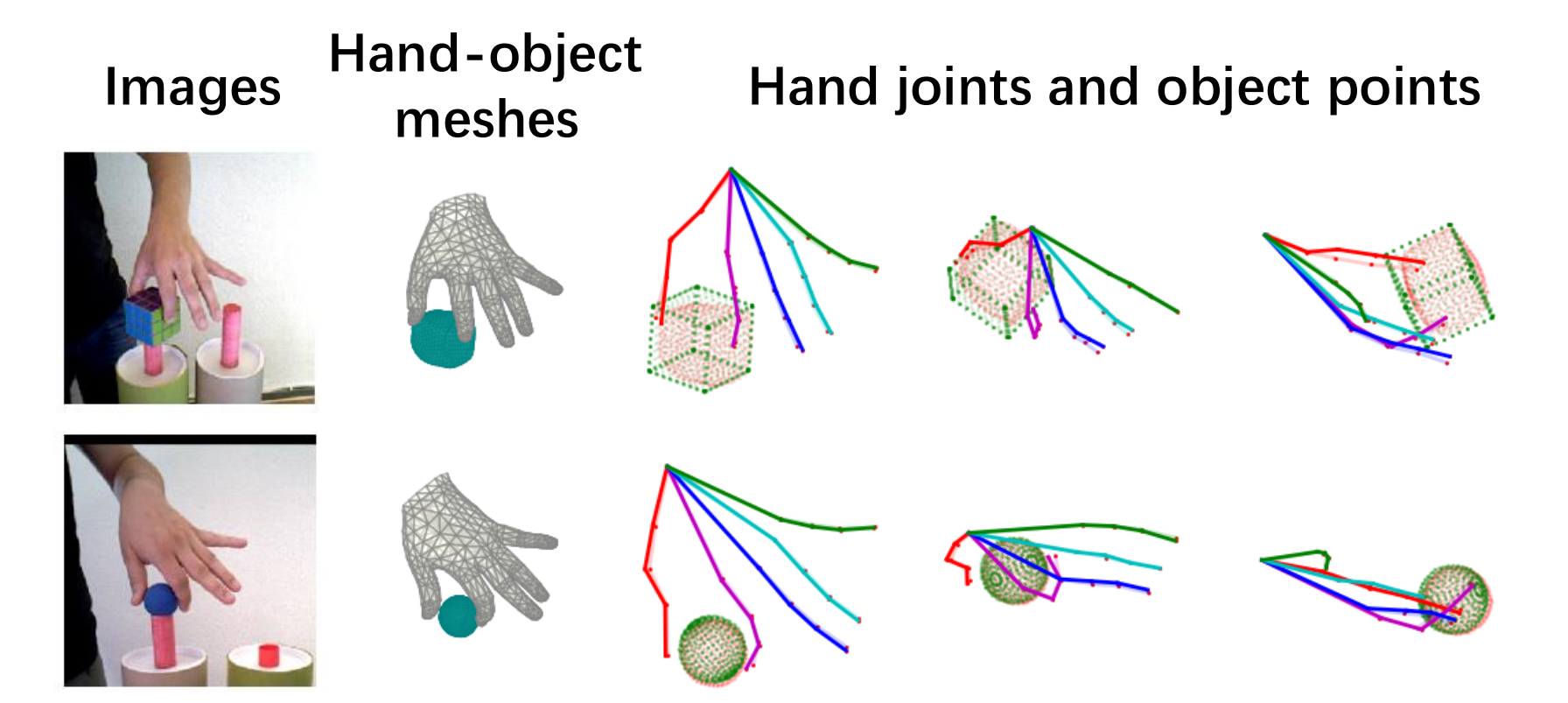}
    \vspace{-5mm}
    \caption{Qualitative visualization of our method on the HIC dataset. We use  the  skeleton  with  opaque  color  to  represent  the estimated  hand  joints  and  the  skeleton  in  semi-transparent  color  to  represent  the  ground  truth.  Besides,  the  green  points  are  sampled  from  the  ground  truth object meshes and the red points are sampled from the estimated object meshes.}
    \label{fig:hic}
    \vspace{-3mm}
\end{figure}

\begin{table}[tb]
\vspace{-3mm}
\caption{Comparison of 3D object reconstruction with \cite{hasson2019learning} on the ObMan synthetic dataset.{CD is the symmetric Chamfer error. PD is the penetration depth.}}
\label{table:cdpd}
\centering
\makebox[0pt][c]{\parbox{0.5\textwidth}
    {
        \centering
        \begin{tabular}{ccc}
    		\hline
    		
    		\hline
    		\centering
    		Method & \quad CD $\downarrow$ \quad & \quad PD  $\downarrow$ \quad\\
    		\hline
    		
    		\hline
    		\cite{hasson2019learning}(w/o contact) & 641.5 & 9.5\\
    		\cite{hasson2019learning}(w/ contact) &637.9&9.2\\
    		\hline
    		Ours(w/o contact) & 405.1&9.6\\
    		Ours(w/ contact) &\textbf{403.8}&\textbf{9.1}\\
    		\hline
    		
    		\hline
    	\end{tabular}
    \vspace{-3mm}
    }}
\end{table}

\begin{table}[tb]
\caption{Comparison with \cite{hasson2019learning} on the HIC dataset.  $^\dagger$ indicates that this value is an approximate value obtained from the histogram.}
\label{table:hic}
\centering
\makebox[0pt][c]{\parbox{0.5\textwidth}
    {
        \centering
        \begin{tabular}{ccc}
    		\hline
    		
    		\hline
    		\centering
    		Method & \quad MPJPE $\downarrow$ \quad & \quad CD $\downarrow$ \quad\\
    		\hline
    		
    		\hline
    		\cite{hasson2019learning} & $27^\dagger$ & $450^\dagger$\\
    		\hline
    		Ours &\textbf{7.5}&\textbf{62.2}\\
    		\hline
    		
    		\hline
    	\end{tabular}
    \vspace{-3mm}
    }
}
\end{table}

\begin{figure}[t]
\centering
	\centering
		\includegraphics[width=0.9\linewidth]{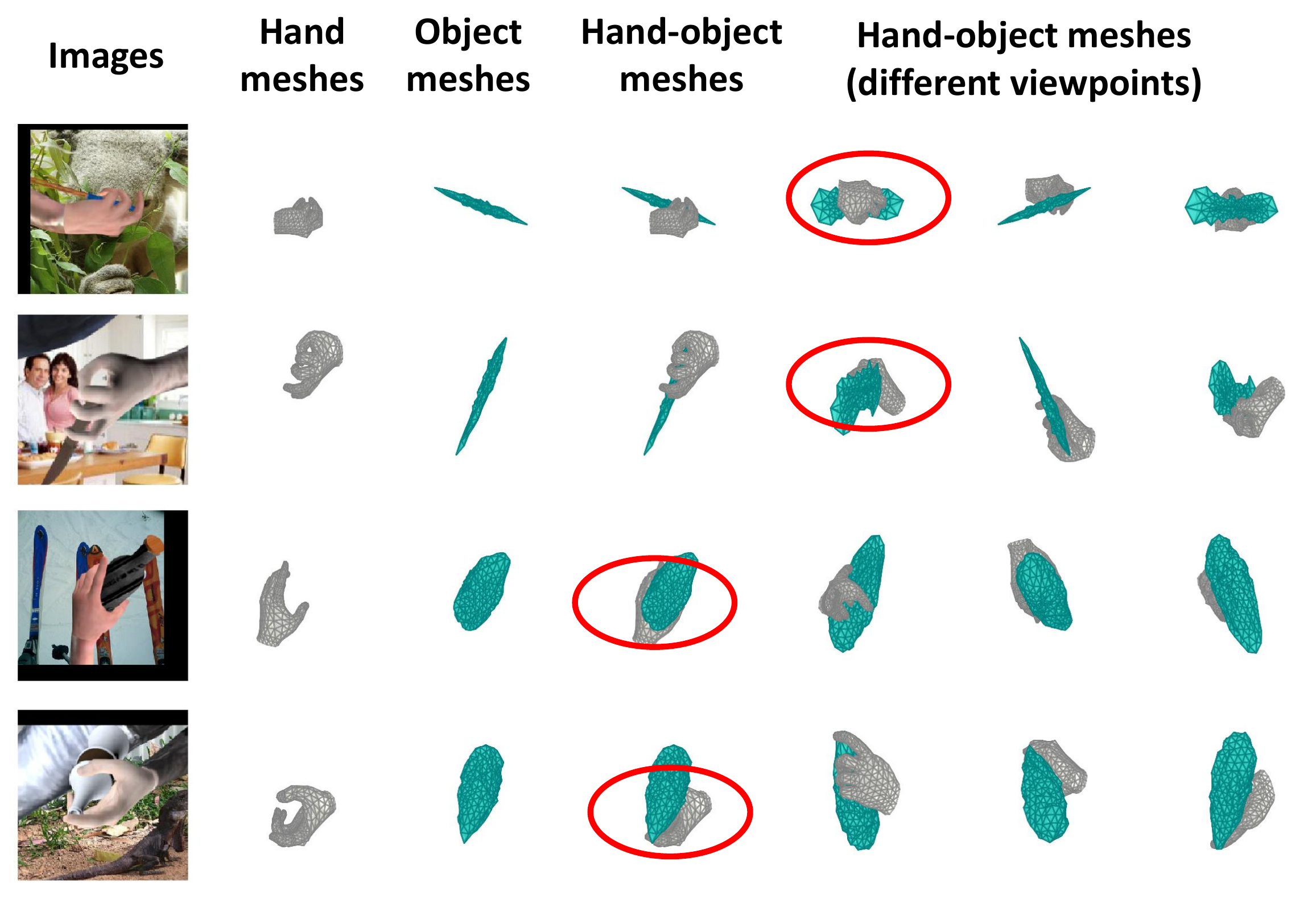}
	~\\
	\vspace{-3mm}
	\caption{Failure cases of our method on the ObMan dataset. From first row to second row, the shape of the invisible part is inaccurate (as circled in red).
	From third row to fourth row, the relative position between hand and object is inaccurate, which usually occurs when fingers covers the object.}
	\vspace{-6mm}
	\label{fig:failure}
\end{figure}

\begin{figure*}[th]
\centering
	\centering
		\includegraphics[width=0.99\linewidth]{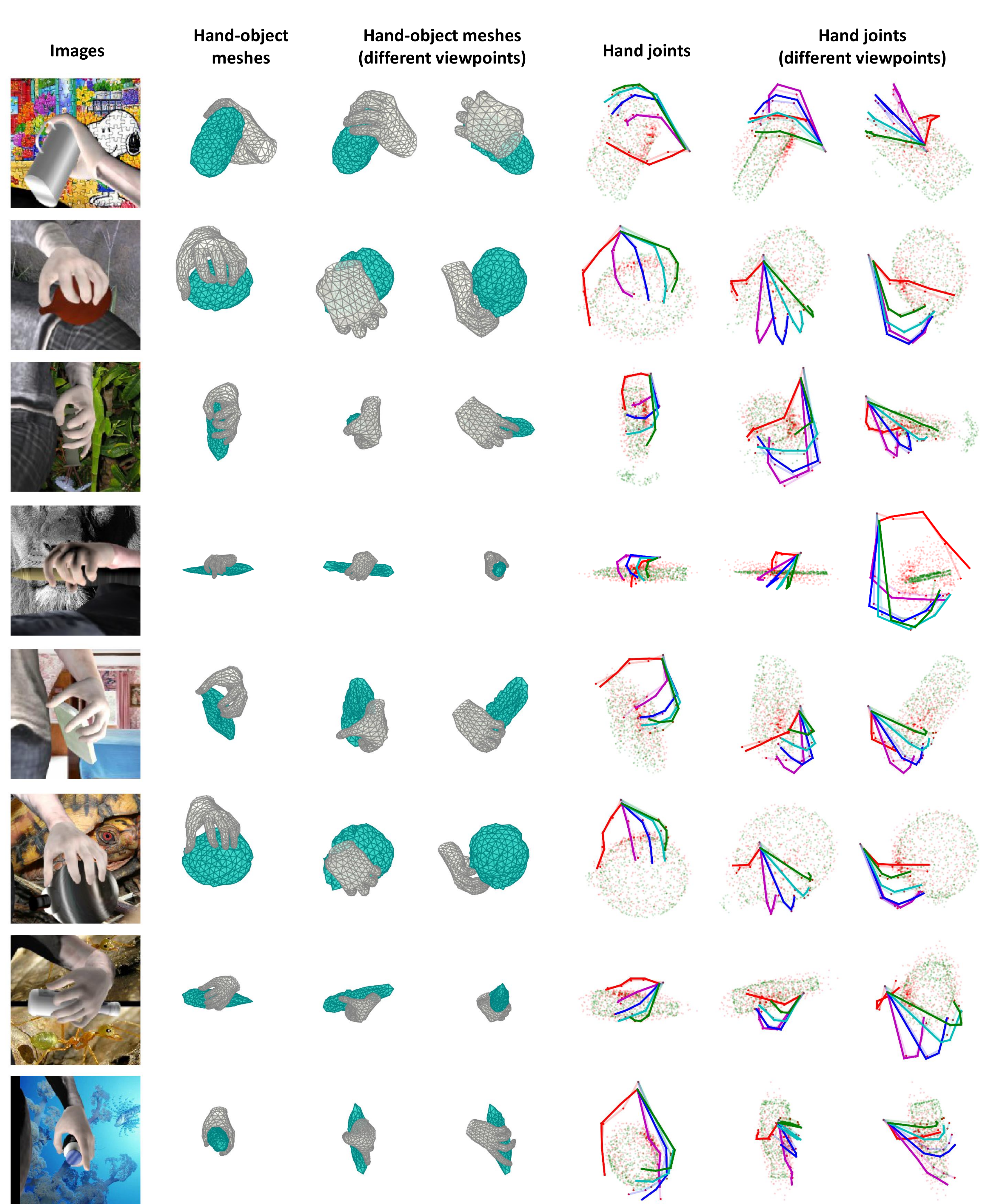}
	\caption{Qualitative visualization of our method on the ObMan dataset. From left to right: input, estimated hand and object meshes (3 columns for 3 different views), estimated 3D hand joints (3 columns for 3 different views). In the columns of hand joints, we use the skeleton with opaque color to represent the estimated hand joints and the skeleton in semi-transparent color to represent the ground truth. Besides, the green points are sampled from the ground truth object meshes and the red points are sampled from the estimated object meshes. Note that scales are changed in the visualization.}
	\label{fig:hoshapes1}
\end{figure*}

\subsubsection{Performance of object estimation}
\label{subsec:object}
The ObMan dataset \cite{hasson2019learning} provides precise object annotations, and we report the object reconstruction accuracy on the ObMan dataset in Table~\ref{table:cdpd}.
When no contact loss is used, the object reconstruction result of ours is already much better than \cite{hasson2019learning}, obtaining a significant error decrease by 36.8\% from 641.5 to 405.1 on CD.
The object CD error further decreases from 405.1 to 403.81 after fine-tuning our network with the contact computation. The object CD error is 36.7\% lower than \cite{hasson2019learning} under the setting with contact computation.
This great improvement in object reconstruction reveals the remarkable effect of our proposed method.

\subsubsection{Results on shape penetration}
\label{subsec:pene}
We report the penetration depth between the hand and object meshes.
In physical space, to obey the rule that hand and object should not share the same space, good shapes of hand and object show low penetration depth.
As shown in Table~\ref{table:cdpd}, the penetration depth of our method is 9.6mm when without the contact computation, and it decreases by 5.2\% and reaches to 9.1mm after adding contact training procedure.
When adding the contact computation, our proposed method gets smaller penetration than \cite{hasson2019learning}.
\subsubsection{Results on the HIC dataset}
\label{subsec:finetune}
Pre-training on the large-scale synthetic data is particularly important for small-scale real datasets.
We fine-tune the model pre-trained on the ObMan dataset to the HIC dataset and compare the performance with \cite{hasson2019learning}.
By using data selection described in Section~\ref{section:id}, the HIC training set consists of only 150 images and the HIC evaluation set consists of 251 images.
The network is pre-trained on the ObMan dataset and then fine-tuned to the HIC training set, and we compare the results of hand-object reconstruction in Table~\ref{table:hic}.
Our method obtains the MPJPE of the hand and the CD of the object in 7.5mm and 62.2mm respectively, while \cite{hasson2019learning} obtains 27mm and 450mm respectively.
These results show that our method has a strong ability to fine-tune the synthetic pre-trained model to real datasets when using a small amount of real-world training data.
We also find that the fine-tuned results show much smaller error than the evaluation results on the large-scale ObMan dataset and the FHBc dataset, We think this may be because there are only two types of objects (i.e., balls and cube) and the configure of hands is similar (i.e., the hand holds the object with two fingers).
This may because there are only two types of objects (i.e., balls and cube) and the configure of hands is similar (i.e., the hand holds the object with two fingers).
Therefore, 150 training samples (with data enhancement via rotation and translation) are sufficient for the proposed network to learn these configurations and obtain good hand-object reconstruction results.

Some visual demonstration images from the test sets of the ObMan and the FHBc are presented in Fig.~\ref{fig:hoshapes}.
The results demonstrate that our method produces high-quality object meshes that can accurately represent the object shape from monocular hand-object images.
The fifth and tenth columns of Fig.~\ref{fig:hoshapes} show that the produced hand and object meshes can reveal hand-object relationships in~3D~space.
More generated shapes from the testing set of ObMan are shown in Fig.~\ref{fig:hoshapes1}.
We also present visualizations from the HIC evaluation set in Fig.~\ref{fig:hic}.

\section{Discussion}
In Fig.~\ref{fig:failure}, we show failure cases, unreasonable object shapes are produced in some unseen parts, and the relative position of the hand and the object is difficult to recover accurately. These problems may be addressed by improving the object representation with more shape priors and reasoning the spatial relationship between the hand and the object.

Shape prior has been proved to be very helpful for 3D reconstruction, both for rigid shape \cite{engelmann2016joint} and non-rigid object \cite{romero2017embodied}.
In our setting, the object tends to be occluded and the category is unknown, which makes the 3D object reconstruction very challenging.
Unlike the category-specific object reconstruction which can use the category information \cite{kanazawa2018learning, Wu_2020_CVPR}, the hand usually interacts with different objects whose shape priors are not shared.
Recent works \cite{tekin2019h+,cao2020reconstructing} estimate object pose and use the pre-defined shape of each object to represent the object surface.
Although the pre-defined shape guarantees the accuracy of the output object, it is impossible to get every object mesh in practical applications.
Without knowing any object category and shape prior, our method (as well as \cite{hasson2019learning}) can acquire a reasonable object surface and reveal the hand-object spatial relationship. Future work will reasonably introduce some shape priors to improve the reconstruction performance.

Reasoning hand-object spatial relationship in 2D or 3D space could also benefit the task.
The mesh-based methods (this work and \cite{hasson2019learning}) use contact loss to learn more reasonable surfaces in hand-object interaction, and
\cite{cao2020reconstructing} optimizes the output hand and object meshes according to their spatial relationship.
Apart from refining the output in 3D space, reasoning the hand-object relationship in 2D could also be helpful.
When the environment is noisy, it is difficult for the network to determine which object is interacting with the hand, it would be more helpful to detect the hand-object interaction in the 2D image space, and then combine the 2D results to perform 3D hand-object reconstruction.

\section{Conclusion}
To handle the challenging task of joint 3D reconstruction of the hand-object image, we present an end-to-end feature fusion network to recover 3D shapes of hand and object.
We introduce a depth estimation module to predict the depth map from the input RGB to enhance the spatial attribute.
To jointly consider the hand and the object branches to extract better features for later reconstruction, we propose to connect the two branches and fuse their latent representations with a feature fusion module.
The proposed LSTM feature fusion module along with several common alternatives have been studied.
We demonstrate that the cross-branch information is helpful for shape reconstruction and that the proposed LSTM feature fusion block is more effective to enhance the object feature after comparing multiple feature fusion strategies.
Experimental results on both synthetic and real-world datasets show that our method outperforms the previous work by a large margin.
As recommended above, future research could explore other cooperation methods and more efficient intermediate representation to achieve better joint reconstruction. We also believe that in addition to investigating cooperation between the hand and the object, research should also be conducted to find cooperation and relationships among more items such as two hands and many objects.

\ifCLASSOPTIONcaptionsoff
  \newpage
\fi

\clearpage



\normalem
\bibliographystyle{IEEEtran}
%

\bibliography{TIP-22941-2020}

\end{document}